\documentclass{article}

\makeatletter
\let\@fnsymbol\@arabic
\makeatother

\usepackage{spconf,amsmath,graphicx}
\usepackage{subfigure}
\usepackage{booktabs,multirow}
\usepackage{amsfonts}
\usepackage{url}
\usepackage{tabularx}
\usepackage{caption}
\usepackage{enumitem}

\graphicspath{{img/}}

\usepackage{xcolor}

\title{ A Comparative Study of Model Adaptation Strategies for Multi-Treatment Uplift Modeling}
\name{Ruyue Zhang$^\dagger$\thanks{$^\dagger$ These authors contributed equally to this work.}, Xiaopeng Ke$^\dagger$\footnotemark[1], Ming Liu$^*$\thanks{$^*$ This work was conducted during his internship at DiDi Chuxing.}, Fangzhou Shi, Chang Men, Zhengdan Zhu}

\address{DiDi Chuxing, Beijing, China}
\begin{document}

%
\maketitle
\begin{abstract}
Uplift modeling has emerged as a crucial technique for individualized treatment effect estimation, particularly in fields such as marketing and healthcare. Modeling uplift effects in multi-treatment scenarios plays a key role in real-world applications. Current techniques for modeling multi-treatment uplift are typically adapted from binary-treatment works. In this paper, we investigate and categorize all current model adaptations into two types: Structure Adaptation and Feature Adaptation. Through our empirical experiments, we find that these two adaptation types cannot maintain effectiveness under various data characteristics (noisy data, mixed with observational data, etc.). To enhance estimation ability and robustness, we propose Orthogonal Function Adaptation (OFA) based on the function approximation theorem. We conduct comprehensive experiments with multiple data characteristics to study the effectiveness and robustness of all model adaptation techniques. Our experimental results demonstrate that our proposed OFA can significantly improve uplift model performance compared to other vanilla adaptation methods and exhibits the highest robustness.
\end{abstract}


\vspace{-10pt}
\section{Introduction}


Recently, Causal Inference has shown its effectiveness in many domains such as online marketing~\cite{sun2023robustness,liu2023explicit,belbahri2021qini}, advertising~\cite{kawanaka2019uplift}, and health care~\cite{liu2024kg}.
Among these, in online marketing, causal effect prediction is vital to subsidy distribution, as subsidies are widely adopted to stimulate user engagement and expand platform user bases.

A key component of causal inference modeling is the causal effect prediction problem with multiple discrete treatment candidates. Recent work on causal effect estimation has mainly focused on binary treatment versus control settings, and extensions are usually required for multi-treatment scenarios. Randomized Controlled Trials (RCTs) are widely regarded as the gold standard for causal inference, serving as the foundation for the development of numerous foundational estimation methods. Among them, representative approaches include tree-based methods~\cite{wager2018estimation}, meta-learners~\cite{kunzel2019metalearners,nie2021quasi}, balancing neural networks (BNN)~\cite{johansson2016learning}, and TARNet~\cite{shalit2017estimating}. Nonetheless, given the ethical and practical limitations of RCTs, a parallel line of work focuses on causal estimation methods that operate on observational data, such as CFRNet~\cite{shalit2017estimating}, DR-CFR~\cite{hassanpour2019learning}, SITE~\cite{yao2018representation}, DragonNet~\cite{shi2019adapting}, and HydraNet~\cite{velasco2023hydranet}. These works mitigate confounding bias through distribution balancing or representation learning. Overall, these methods have demonstrated strong performance in binary treatment settings.

However, how to extend binary treatment models to multi-treatment scenarios remains an open question—unfortunately, current research in this area lacks systematic and reliable investigation. In multi-treatment uplift modeling, binary models are often extended in different ways depending on how the treatment variable is incorporated into the model structure. Particularly for modeling with observational data, the adjustment for distributional differences also needs to be adapted to the multi-treatment setting. Methods such as tree-based models and BNN typically \textbf{treat the treatment as an input feature, which is a broadly applicable approach}. In contrast, for binary models with a two-head structure such as CFRNet, \textbf{a common practice is to extend the architecture into a multi-head form}. The MEMENTO~\cite{mondal2022memento} model exemplifies this approach, with the main difference being its specific method for balancing data distribution. Nonetheless, neither of these approaches consistently delivers stable performance across different datasets, underscoring the limitations of current extension strategies and calling for more efficient and flexible solutions for multi-treatment uplift modeling.

To figure out the robust modeling method for multi-treatment scenarios, we organize all modeling approaches for multi-treatment scenarios and classify them into two groups: \textbf{Structural Adaptation} and \textbf{Feature Adaptation}. Additionally, we introduce a novel multi-treatment modeling approach, \textbf{Orthogonal Function Adaptation}, which is designed as a plug-and-play module that enhances model generalization across diverse datasets. Its architecture consistently improves estimation performance under various data conditions with minimal additional computational overhead. Our experimental evaluation encompasses a diverse collection of datasets, including randomized controlled trials (RCTs), observational studies, synthetically generated data with noise, and data satisfying monotonicity constraints. Results demonstrate that the proposed OFA method consistently outperforms existing baselines across the majority of these settings.


Our main contributions are summarized as follows.
\begin{itemize}[leftmargin=*]
\item To the best of our knowledge, we are the first to categorize current works on multi-treatment uplift modeling and conduct a comprehensive study on this scenario.
\item We propose the Orthogonal Function Adaptation (OFA) technique for multi-treatment modeling based on the Weierstrass approximation theorem. This method is plug-and-play and can maintain computational complexity while increasing the treatment options.
\item We conduct comprehensive experiments on synthetic and real-world datasets. The experimental results demonstrate that the proposed OFA outperforms other vanilla adaptations across various data characteristics.
\end{itemize}


\section{Methodology}
\label{sec:2}

\begin{figure}
    \centering
    \includegraphics[width=\linewidth]{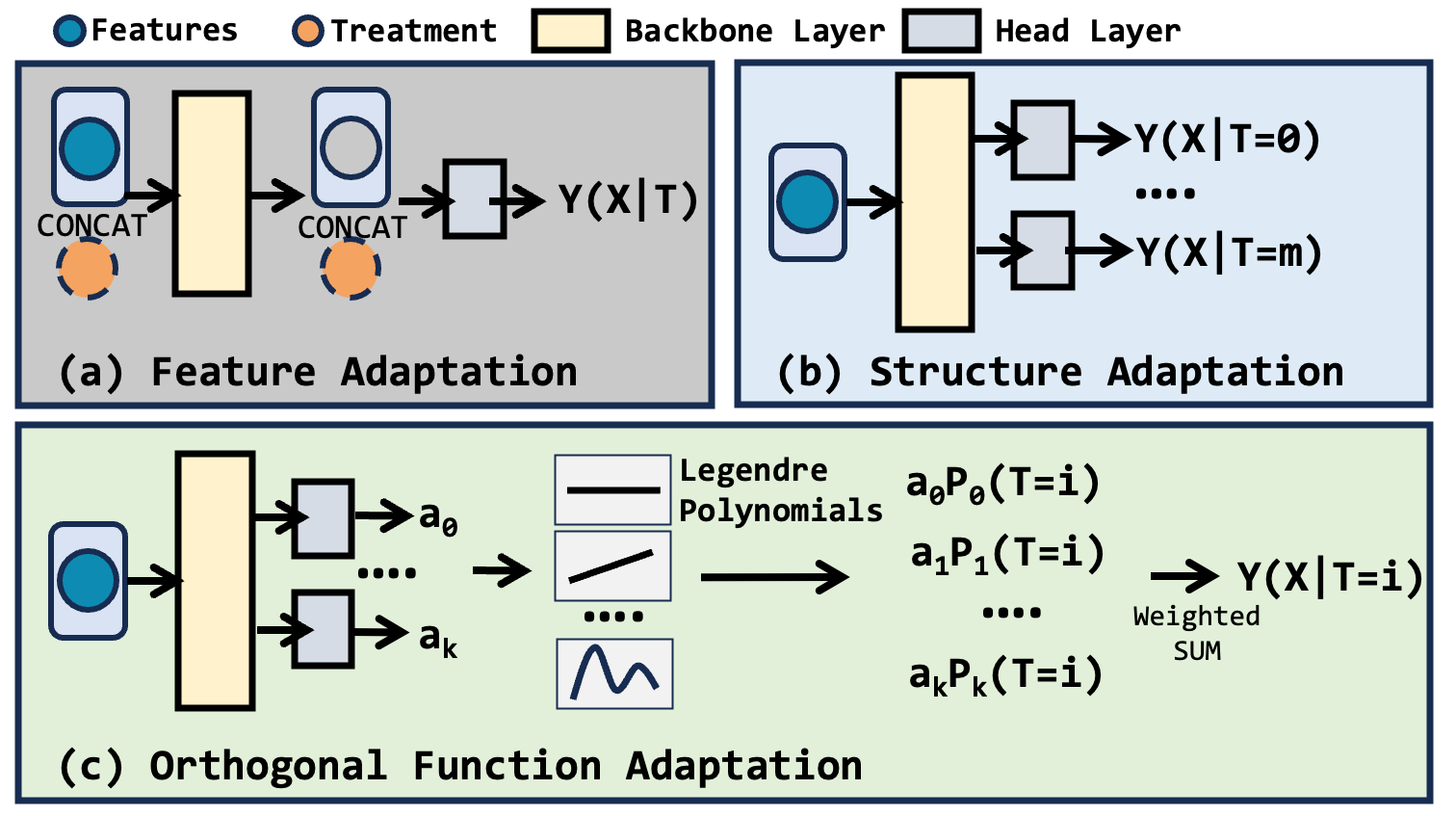}
    \caption{Overview of different model adaptation categories}
    \label{fig:model-adaptation-overview}
    \vspace{-5pt}
\end{figure}

\subsection{Notation}
Let $\mathcal{D} = \{X_i,Y_i,T_i\}_{i=1}^{N}$ denotes the dataset where $X_i\in \mathbb{R}^d$ represents the $d$-dimensional input feature vector, $Y_i$ denotes the label, and $T_i\in \mathcal{T}$ denotes the treatment. $\mathcal{T} = \{t_1,t_2,\cdots,t_m\}$ is the set of $m$ possible treatment values. In this paper, we discuss the adaptation for multi-valued treatments under the Neyman-Rubin potential outcome framework~\cite{rubin2005causal}. We assume the uplift model has the following form:
\begin{equation}
    f(X,T;\theta_f) = g(\varphi_{X};T) \nonumber
\end{equation}
where $\varphi_{X}$ denotes the hidden feature, and $g(\cdot,\cdot)$ denotes the associated effect estimator that operates on the hidden feature.

\subsection{Feature Adaptation}



Feature Adaptation technique is to concatenate the treatment value with the features as shown in Figure~\ref{fig:model-adaptation-overview}(a).
The corresponding model can be written as
\begin{equation}
    f_{FA}(X,T;\theta_{f_{FA}}) = g(\text{concat}(\varphi_X, T)) \nonumber
\end{equation}
where $\text{concat}(\cdot,\cdot)$ denotes the concatenation function.

\subsection{Structure Adaptation}

To handle multi-valued treatments, another common way is to construct branches from the original model as shown in Figure~\ref{fig:model-adaptation-overview}(b). This can be represented as the set of effect estimators $\mathcal{G} = \{g_1, g_2, \ldots, g_m\}$. Consequently, the model becomes
\begin{equation}
    f_{SA}(X,T;\theta_{f_{SA}}) = g_{T}(\varphi_X). \nonumber
\end{equation}

Note that the hidden feature $\varphi_X$ could be the input feature $\varphi_X=X$, and the whole model degenerates into several separate submodels.

\subsection{Orthogonal Function Adaptation}

Through the Weierstrass approximation theorem~\cite{stone1948generalized}, we can use a set of orthogonal polynomials to approximate every continuous function. Here, we propose the Orthogonal Function Adaptation (OFA) method as shown in Figure~\ref{fig:model-adaptation-overview}(c). We modify the model to generate coefficients for each polynomial function. Without loss of generality, we utilize orthogonal polynomials~\cite{szeg1939orthogonal} to estimate the target function. The model can be written as
\begin{equation}
   f_{OFA}(X,T;\theta_{OFA}) = \sum_{j=0}^{p} a_j(\varphi_X)\cdot P^{(j)}(T) \nonumber  
\end{equation}
where $p$ is the highest degree of the polynomial, $a_j(\cdot)$ denotes the coefficient estimator, and $P^{(j)}$ is the $j$-th orthogonal polynomial. In this paper, we select the Legendre orthogonal polynomial~\cite{weisstein2002legendre} to construct the approximation. These polynomials can be written as
\begin{align}
    P^{(0)}(t)&=1, P^{(1)}(t)=t, \nonumber\\
    (k+1)P^{(k+1)}(t)&=(2k+1)tP^{(k)}(t)-kP^{(k-1)}(t).\nonumber
\end{align}

Using the theorem from \cite{galanti2020modularity}, we can also prove that the OFA achieves the lowest complexity while the error $\leq\varepsilon$:
\begin{equation}
\small
    C_{\text{OFA}} = \mathcal{O}(\varepsilon^{-\frac{m}{r}}) < C_{\text{SA}}=\mathcal{O}(\varepsilon^{-\frac{d}{r}}) < C_{\text{FA}}=\Omega(\varepsilon^{-\frac{m+d}{r}})
    \nonumber
\end{equation}
where $C$ is the model complexity and $m < d$ must hold.

\subsection{Loss Design}
We select the Binary Cross-Entropy Loss as the basic loss. It could be written as:
\begin{equation}
\small
    \mathcal{L}_{\text{BCE}} = \frac{1}{N}\sum_{i=1}^N Y_i\log(f(X_i,T_i;\theta)) + (1-Y_i)\log(1-f(X_i,T_i;\theta))\nonumber
\end{equation}
As for handling the selection bias of the observation data, we apply the balanced representation technique in this paper. We select the Maximum Mean Discrepancy (MMD)~\cite{gretton2012kernel} and Wasserstein discrepancy (WASS)~\cite{vallender1974calculation} function to measure the distance between two feature distributions. This discrepancy loss can be written as
\begin{equation}
    \mathcal{L}_{\text{disc}} = \text{disc}(\{h(X_i;T_i=t_p)\}_{i=1}^N, \{h(X_i;T_i=t_q)\}_{i=1}^N) \nonumber
\end{equation}

where $ t_p\neq t_q$ and $1\leq p,q\leq m$. Thus, our final loss can be written as:
\begin{equation}
    \mathcal{L} = \lambda_1 \cdot \mathcal{L}_{\text{BCE}} + \lambda_2\cdot \mathcal{L}_{\text{disc}} \nonumber
\end{equation}
where $\lambda_1,\lambda_2\in \mathbb{R}$ are the coefficients.

\section{Experiments}
\label{sec:3}

In this section, we conduct experiments to answer the following research questions.
\begin{itemize}[leftmargin=*]
    \item \textbf{RQ1: How does different adaptations perform on all synthetic and real-world datasets?}
    \item \textbf{RQ2: Is our OFA more robust than other adaptations?}
    \item \textbf{RQ3: Does the OFA maintain the improvement when applied to different models?}
\end{itemize}

\subsection{Datasets and Implementation Details}

\textbf{Synthetic Dataset.} 
We construct synthetic datasets with five treatment candidates. Each sample consists of eight covariates, one treatment variable, and one outcome, with 10,000 samples for training and 10,000 for testing; the test set is always a noise-free RCT. We consider three settings: (1) RCT data where treatments are randomized with small noise, including two variants: RCT-Noise, which injects additional noise into outcomes, and RCT-NM, which employs complex functional forms and removes monotonicity constraints on ITE curves; (2) observational data where treatment assignment depends on covariates, with or without an instrumental variable; and (3) mixed data that combines RCT and observational samples in a 1:1 ratio, also with or without an instrumental variable.

\textbf{Real-World Dataset.} 
We collect a real-world dataset named RW-RCT from a ride-hailing platform's coupon experiment. Each sample contains 47 covariates and five treatment candidates. The RW-RCT training set contains 975,940 samples; the corresponding test set consists of 811,027 samples.

\textbf{Hyper-Parameter Setting.}
Across both synthetic and real-world datasets, the number of parameters in all models is kept comparable, with around 90k parameters for synthetic datasets and about 150k for real-world datasets due to their larger scale. All models underwent multiple rounds of hyper-parameter search, and each selected configuration was evaluated through repeated independent runs to ensure that the results reliably reflect the models' performance. For all experiments, we used the Adam optimizer with a fixed learning rate of 1e-4.

\subsection{Evaluation Metrics}

In this paper, we use the mean of the Qini score~\cite{belbahri2021qini} to measure the performance of the uplift models. This metric can be written as
\begin{align}
     \text{mQini} &= \frac{1}{m}\sum_{i=1}^m \text{Qini}(f,t_i,\mathcal{D}) \nonumber\\
     &=\frac{1}{m}\sum_{i=1}^m\int\Big[ Y_T(k) - \frac{N^T(k)}{N_C(k)}Y_C(k) \Big]_{f,t_i,\mathcal{D}}dk \nonumber
\end{align}

\begin{table}[t]
    \centering
    \begin{tabularx}{0.48\textwidth}{@{}cccc@{}}
    \toprule
    \textbf{Method} 
    & \textbf{RCT} 
    & \textbf{RCT-Noise}
    & \textbf{RCT-NM} \\
    \midrule
    Slearner + FA & \underline{0.2854} & 0.2552 & 0.2488 \\
    BNN + FA & \textbf{0.2855}  & 0.2663& 0.2812\\
    TARNet + SA & 0.1977 & 0.2442 & 0.1948 \\
    DR-CFR + SA & 0.2023 & {0.2455} & {0.2024} \\
    \midrule
    TARNet + OFA & 0.2547 & \underline{0.2732} & \textbf{0.3160}\\
    DR-CFR + OFA & 0.2652 & \textbf{0.2875} & \underline{0.2882} \\
    \bottomrule
    \end{tabularx}
    \caption{mQini Score on three synthetic RCT datasets.}
    \label{tab:rct-perf}
    \vspace{-10pt}
\end{table}

\subsection{Model Performance (RQ1)}

On synthetic datasets, OFA consistently shows stable and measurable improvements. Table~\ref{tab:rct-perf} reports the results on RCT data. Under near-ideal low-noise conditions, FA models such as BNN+FA achieved the highest scores, since their flexibility is maximized in this setting. Nevertheless, even in such minimally confounded cases, OFA still achieved about a 30.0\% improvement over SA. When noise was introduced, the advantage of OFA became more evident, yielding an 8.0\% improvement over the best non-OFA model, BNN+FA. On RCT-NM data, where monotonicity constraints were removed and response curves exhibited more complex functional forms, OFA maintained its superiority, achieving a 12.4\% gain over the best non-OFA model. Furthermore, Tables~\ref{tab:obs-perf} and \ref{tab:mix-perf} present the results on observational and mixed datasets. Regardless of whether instrumental variables were included, OFA-based models consistently ranked at the top. On purely observational data with instrumental variables, the best OFA model improved by 42.1\% compared with the best non-OFA model, and by 41.2\% without instrumental variables. On mixed datasets, the improvements were 13.7\% and 14.1\% with and without instrumental variables, respectively.  

Table~\ref{tab:real-world-perf} further demonstrates the results on real-world RCT data. TARNet enhanced with OFA achieved the best performance, with a 6.6\% improvement over the best non-OFA model, providing strong evidence of its effectiveness. This result indicates that the robustness demonstrated by OFA on synthetic datasets can directly transfer to real-world applications, leading to superior practical outcomes.  


The results demonstrate that although FA can reach peak performance under ideal, low-noise environments, OFA provides substantial, reliable improvements over both SA and FA across all other scenarios, with its advantage particularly pronounced in noisy and complex settings. Furthermore, this robustness makes OFA the best-performing method overall on real-world business RCTs. Across the entire spectrum of datasets, OFA establishes itself as the most reliable and consistently high-performing approach, delivering significant performance gains in the most challenging scenarios.

\begin{table}[t]
    \centering
    \begin{tabularx}{0.48\textwidth}{@{}ccc@{}}
    \toprule
    \textbf{Method} 
    & \textbf{OBS w/ IV} 
    & \textbf{OBS w/o IV} \\
    \midrule
    BNN + FA + WASS & {0.1249}  & 0.1014\\
    BNN + FA + MMD & {0.0860}  & 0.0791\\
    CFRNet + SA + WASS & 0.1997 & 0.1843  \\
    CFRNet + SA + MMD & 0.1795 & 0.1718 \\
    DR-CFR + SA + WASS & {0.1747} & 0.2017 \\
    DR-CFR + SA + MMD & {0.1721} & 0.1740 \\
    \midrule
    CFRNet + OFA & {0.2778} & {0.2662} \\
    CFRNet + OFA + WASS & \underline{0.2796} & {0.2818} \\
    CFRNet + OFA + MMD & 0.2789 & \underline{0.2826} \\
    DR-CFR + OFA + WASS & 0.2652 & \textbf{0.2848} \\
    DR-CFR + OFA + MMD & \textbf{0.2837} & {0.2797} \\
    \bottomrule
    \end{tabularx}
    \caption{mQini Score on synthetic observation datasets (with or without instrument variable).}
    \label{tab:obs-perf}
    \vspace{-5pt}
\end{table}

\begin{table}[t]
    \centering
    \begin{tabularx}{0.48\textwidth}{@{}ccc@{}}
    \toprule
    \textbf{Method} 
    & \textbf{MIX w/ IV} 
    & \textbf{MIX w/o IV} \\
    \midrule
    BNN + FA + WASS & {0.2345}  & 0.2510\\
    BNN + FA + MMD & {0.2460}  & 0.2357\\
    CFRNet + SA + WASS & 0.1892 & 0.2363  \\
    CFRNet + SA + MMD & 0.2007 & 0.1971 \\
    DR-CFR + SA + WASS & {0.1863} & 0.2019 \\
    DR-CFR + SA + MMD & {0.2025} & 0.1714 \\
    \midrule
    CFRNet + OFA & {0.2522} & {0.2595} \\
    CFRNet + OFA + WASS & {0.2684} & {0.2744} \\
    CFRNet + OFA + MMD & 0.2674 & \textbf{0.2864} \\
    DR-CFR + OFA + WASS & \underline{0.2730} & {0.2769} \\
    DR-CFR + OFA + MMD & \textbf{0.2798} & \underline{0.2776} \\
    \bottomrule
    \end{tabularx}
    \caption{mQini Score on synthetic mixed datasets (with or without instrument variable).}
    \vspace{-5pt}
    \label{tab:mix-perf}
\end{table}


\subsection{Robustness Analysis (RQ2)}
As shown in Table~\ref{tab:rct-perf}, under noisy RCT conditions, OFA provides inherent regularization through its orthogonal basis functions, effectively resisting perturbations and yielding an 8.0\% improvement over the best non-OFA method. When the response curve follows complex, non-monotonic patterns, OFA exhibits strong adaptability, as evidenced by a 62.2\% improvement of TARNet + OFA over its SA counterpart. In observational scenarios (Table~\ref{tab:obs-perf} to \ref{tab:mix-perf}), OFA combined with representation learning offers a stable prediction layer that enhances generalization from biased samples, resulting in more reliable bias correction.

Based on comprehensive experimental results, our proposed OFA method demonstrates significantly superior robustness compared to other adaptation approaches, with quantified advantages across three key dimensions: noise resilience, functional flexibility, and bias correction stability. 



\subsection{Transferability (RQ3)}
As shown in Tables~\ref{tab:rct-perf} to \ref{tab:real-world-perf}, our proposed OFA consistently improves performance across different model architectures.



When applied to TARNet, OFA enhanced its performance across all challenging scenarios. Most notably, on the RCT-NM dataset, TARNet + OFA achieved a massive 62.2\% improvement over the vanilla TARNet + SA. Similarly, when plugged into DR-CFR, the OFA variant also became the top performer on the noisy RCT dataset, showcasing a 17.1\% improvement over its standard counterpart.


In conclusion, OFA is not only effective for a single model but acts as a versatile and plug-and-play performance enhancer. Its consistent ability to elevate the performance of diverse underlying architectures confirms its strong generalizability and establishes it as a universally valuable component for multi-treatment causal effect estimation.

\begin{table}[t]
    \centering
    \begin{tabularx}{0.3\textwidth}{@{}cc@{}}
    \toprule
    \textbf{Method} 
    & \textbf{RW-RCT }
     \\
    \midrule
    Slearner + FA  & {0.1082} \\
    BNN + FA  & {0.0880} \\
    TARNet + SA & 0.0765  \\
    DR-CFR + SA & {0.0896} \\
    \midrule
    TARNet + OFA & \textbf{0.1153} \\
    DR-CFR + OFA & \underline{0.1100} \\
    \bottomrule
    \end{tabularx}
    \caption{mQini Score on the real-world RCT dataset.}
    \label{tab:real-world-perf}
    \vspace{-4pt}
\end{table}

\section{Conclusion}
\label{sec:4}

In this paper, we categorize and study different model adaptations for multi-treatment scenarios. We further propose the Orthogonal Function Adaptation (OFA) method, which has superior robustness and generalizability across diverse datasets and baselines. It significantly enhances estimation accuracy under noise, selection bias, and complex functional relationships, while presenting  stable performance gains in both synthetic and real-world datasets.

\vfill\pagebreak



\bibliographystyle{IEEEbib}
\bibliography{refs}

\end{document}